\def\year{2022}\relax
\documentclass[letterpaper]{article} 
\usepackage{aaai22}  
\usepackage{times}  
\usepackage{helvet}  
\usepackage{courier}  
\usepackage[hyphens]{url}  
\usepackage{graphicx} 
\usepackage{natbib}  
\usepackage{caption} 
\usepackage{multirow}
\usepackage{subcaption}
\usepackage{amsmath}
\usepackage{amssymb}
\DeclareCaptionStyle{ruled}{labelfont=normalfont,labelsep=colon,strut=off} 
\usepackage{algorithm}
\usepackage{algorithmicx}
\usepackage[noend]{algpseudocode}

\usepackage[dvipsnames]{xcolor}
\usepackage{newfloat}
\usepackage{listings}
\floatstyle{ruled}
\newfloat{listing}{tb}{lst}{}
\floatname{listing}{Listing}
\title{Powering Finetuning in Few-Shot Learning: Domain-Agnostic Bias Reduction with Selected Sampling}

\author {
    Ran Tao,
    Han Zhang, 
    Yutong Zheng, 
    Marios Savvides 
}
\affiliations {
    Carnegie Mellon University\\
    taoran1@cmu.edu, {Hanz3, yutongzh, marioss}@andrew.cmu.edu}

\usepackage{bibentry}

\begin{document}
\maketitle

\begin{abstract}
In recent works, utilizing a deep network trained on meta-training set serves as a strong baseline in few-shot learning. In this paper, we move forward to refine novel-class features by finetuning a trained deep network. Finetuning is designed to focus on reducing biases in novel-class feature distributions, which we define as two aspects: class-agnostic and class-specific biases. Class-agnostic bias is defined as the distribution shifting introduced by domain difference, which we propose Distribution Calibration Module(DCM) to reduce. DCM owes good property of eliminating domain difference and fast feature adaptation during optimization. Class-specific bias is defined as the biased estimation using a few samples in novel classes, which we propose Selected Sampling(SS) to reduce. Without inferring the actual class distribution, SS is designed by running sampling using proposal distributions around support-set samples. By powering finetuning with DCM and SS, we achieve state-of-the-art results on Meta-Dataset with consistent performance boosts over ten datasets from different domains. We believe our simple yet effective method demonstrates its possibility to be applied on practical few-shot applications.

\end{abstract}

\begin{figure}[t]
     \centering
     \begin{subfigure}{0.9\linewidth}
         \centering
         \includegraphics[width=\linewidth]{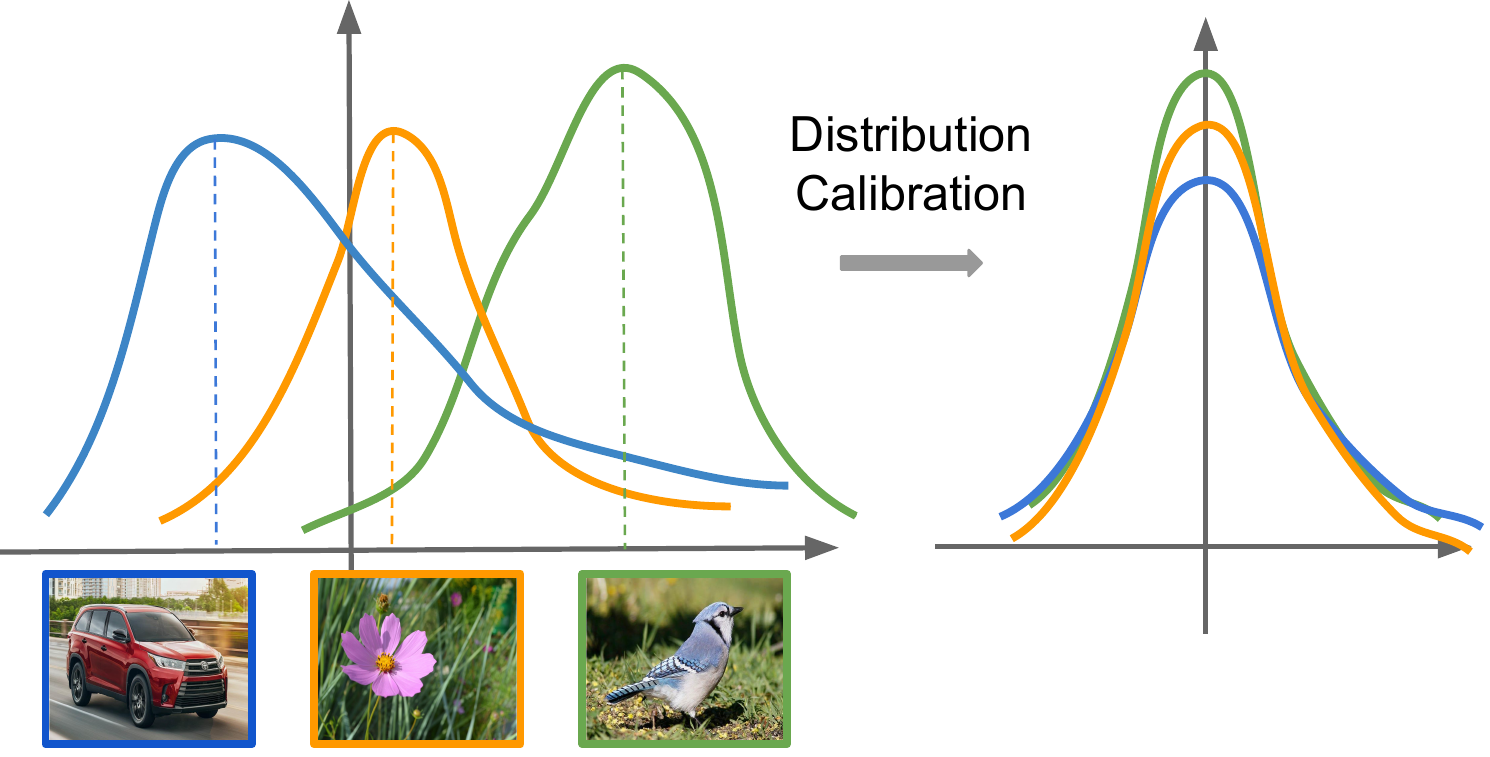}
         \caption{Reducing Class-agnostic Bias by Distribution Calibration Module(DCM). For novel classes from different domains, the feature distribution could be skewed and our proposed DCM is to eliminate the domain difference in feature distribution. }
         \label{fig:DCM}
     \end{subfigure}
     \begin{subfigure}{0.9\linewidth}
         \centering
         \includegraphics[width=\linewidth]{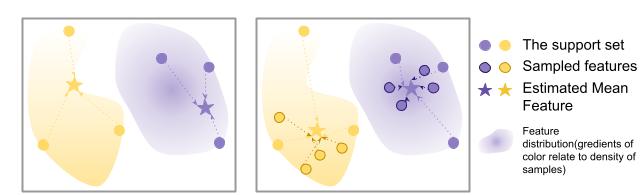}
         \caption{Reducing Class-specific Bias by Selected Sampling. As shown in the left, a few samples leads to a biased mean estimation. In the right, by selectively sampling more features, the bias in estimation is largely reduced. The mean estimation is rectified towards the area with higher sample density. }
         \label{fig:ss}
     \end{subfigure}
    \caption{Illustration on Biases in Feature Distribution. 
    }
    \label{fig: main}
    \vspace{-0.5cm}
\end{figure}
\vspace{-0.2cm}

\section{Introduction}

In recent works \cite{chen2019closer, tian2020rethinking, chen2020new, dhillon2019baseline}, the importance of utilizing a good feature embedding in few-shot learning is well studied and addressed. A feature embedding is pre-trained as a classification task using meta-training set(base classes). Finetuning on the meta-test set(novel classes) \cite{tian2020rethinking, yang2021free, dhillon2019baseline} is shown to surpass most meta-learning methods. 
However, only finetuning a classifier on the meta-test set leaves the feature embedding unchanged. A pre-trained feature extractor is sufficient to have well-defined feature distributions on base classes, while this is not true for novel classes.
Novel classes may come from a variety of domains different from base classes. 
Globally, initial feature distributions of novel classes could be affected mainly by the domain difference. Locally, features are not trained to cluster tightly within a class and separate well between classes, which intensifies the biased estimation of only a few samples. Those biases in novel-class feature distributions address the importance of refining novel-class features. 

In our work, we refine novel-class features by finetuning the feature extractor on the meta-test set using only a few samples. We focus on reducing biases in novel-class feature distributions by defining them into two aspects: class-agnostic and class-specific biases. 
Class-agnostic bias refers to the feature distribution shifting caused by domain differences between novel and base classes. The unrefined features from novel classes could cluster in some primary direction due to the domain difference, which leads to the skewed feature distribution as shown in Fig.~\ref{fig:DCM}. In other words, the feature distribution of novel classes is shifted by the domain difference when directly using the pre-trained feature extractor. 
Class-specific bias refers to the biased estimation using only a few samples in one class. Biased estimation is always critical for few-shot learning. By only knowing a few samples, the estimation of feature distribution within one class is biased as shown in Fig.~\ref{fig:ss}. The bias between empirical estimation with its true value would be reduced with more samples involved. Running sampling under each class distribution is the most direct way to enlarge the support set. However, this is not applicable when each class distribution is unknown. 

In our work, we propose the Distribution Calibration Module(DCM) to reduce class-agnostic bias. DCM is designed to eliminate domain difference by normalizing the overall feature distribution for novel classes and further reshaping the feature manifold for fast adaptation during finetuning.
For class-specific bias, we propose Selected Sampling(SS) to augment more data for better estimation. More specifically, the Selected Sampling happens on a chain of proposal distributions centered with each data point from the support set. 
And the whole sampling process is guided by an acceptance criterion that only samples beneficial to the optimization will be selected. By eliminating the domain difference through distribution calibration, DCM boosts performance over ten datasets from different domains on Meta-Dataset evaluation, demonstrating the importance of reducing class-agnostic bias when dealing with domain issues. Meanwhile, we theoretically analyze how DCM is designed for fast feature adaptation and showcase its supreme faster convergence compared with direct finetuning in Fig.~\ref{fig:convergence}. Furthermore, based on DCM's competitive performance, reducing class-specific bias with Selected Sampling can further enhance the finetuning performance by a large margin over ten datasets. Without inferring the actual feature distribution, our selected sampling can effectively explore unknown feature space and augment features to reduce class-specific bias in estimation. 

\textbf{Our contributions}:
\textbf{1)} We address the understanding of biases in novel-class feature distributions when using a pre-trained feature extractor. By proposing to reduce class-agnostic and class-specific biases through DCM and SS, we power finetuning with domain-agnostic consistent performance gain.
\textbf{2)} We propose an efficient Selected Sampling strategy to direct sample features for class-specific bias reduction. Without inferring the class distribution, the selected sampling effectively enlarges the support set with informative data on the feature space. 
\textbf{3)} We evaluate our method by performing comprehensive experiments on Meta-Dataset. We achieve State-of-the-Art performance and a remarkably consistent performance improvement over ten datasets from different domains. We hope this work could contribute to the understanding of the feature space for classification-oriented tasks as well.

\section{Related Work}

\subsection{Overview of the Few-Shot Problem}

Few-shot learning has been quite an active research field in recent years.
The branch of Meta-learning methods \cite{finn2017model, rusu2018meta, vinyals2016matching, snell2017prototypical, sung2018learning, chen2020new, simon2020adaptive} on few-shot learning is designed to directly back-propagate the loss of the test set while the hypothesis for classification is proposed with the training set. 
Also methods are not adherent to meta-learning only. There are: data argumentation with hallucinating more samples\cite{hariharan2017low, wang2018low}, optimization with ridge regression or support vector machine \cite{bertinetto2018meta, lee2019meta}, using graph neural networks \cite{garcia2017few, kim2019edge}, self/semi-supervised learning \cite{ren2018meta, gidaris2019, li2019learning, wang2020instance}, learning with semantic information \cite{li2020boosting}, class weight generalization \cite{gidaris2018dynamic, gidaris2019generating, guo2020attentive}, modules working on attentive spatial features \cite{li2019finding, hou2019cross, doersch2020crosstransformers}, knowledge distillation \cite{tian2020rethinking}. The recent \cite{triantafillou2019meta} proposes a more realistic evaluation for few-shot learning where algorithms are evaluated over 10 datasets from different domains with a large-scale meta-training set spanned from ImageNet\cite{krizhevsky2012imagenet}. The evaluation on meta-Dataset not only requires algorithms to obtain a good performance on few-shot learning but also sets higher demands on generalization over different domains. 

Distribution calibration \cite{yang2020dpgn, yang2021free} raises attention in few-shot learning recently. In \cite{yang2021free}, to conduct sampling, the similarities between base and novel classes are used to transfer the distribution parameters from base classes to novel classes. The similarity measurement may limit its application for cross-domain problems. 
Unlike their method, we tackle the distribution calibration from coarse to refined bias reduction. 
The class-agnostic bias reduction is conducted to eliminate skewness shown in the overall feature distribution. Further, class-specific bias is reduced by sampling using proposal distribution centered with the support set. 
Our methods successfully run direct sampling on the feature space without inferring the actual feature distribution for each class. 
There have been some works on feature transformations in few-shot learning recently. \cite{wang2019simpleshot} uses the mean estimation of base classes to normalize novel-class features, without considering the domain difference between novel and base classes. \cite{liu2020embarrassingly} proposes random pruning features after normalization for one-shot learning, and the random pruning is expected to find part of feature embedding that fits in the domain of novel classes. We directly solve the domain shifting issue in feature distribution, and as illustrated later, those class-sensitive features are further amplified during finetuning.
Finetuning results in \cite{triantafillou2019meta} is to finetune the K-Nearest Neighbor trained model with the meta-training set(followed by inner-loop finetuning on meta-test set); the experiments with our methods are more precisely defined as \textit{(inner-loop) finetuning}, which is only to use the support set within one episode from meta-test set during evaluation. \textit{(Inner-loop) finetuning} is an attempt to reach a good classification through only a handful of training samples. \cite{dhillon2019baseline} proposes transductive finetuning which involves the query set also.

\section{Method} \label{sec:method}
\subsection{Leverage a Feature Extractor to Few-Shot Problem}

We first formalize the few-shot classification setting with notation. Let $(\mathbf{x}, y)$ denote an image with its ground-truth label. In few-shot learning, training and test sets are referred to as the support and query set respectively and are collectively called a $C$-way $K$-shot episode. 
We denote training(support) set as $\mathcal{D}_s = \{(\mathbf{x_i}, y_i)\}_{i=1}^{N_s}$ and test(query) set as $\mathcal{D}_q = \{(\mathbf{x_i}, y_i)\}_{i=1}^{N_q}$, where $y_i \in C$ and $|C|$ is the number of ways or classes and $N_s$ equals to $C \times K $. 

For supervised learning, a statistics $\theta^* = \theta^*(\mathcal{D}_s)$ is learnt to classify $\mathcal{D}_s$ as measured by the cross-entropy loss:

\begin{equation}
    \theta^*(\mathcal{D}_s) = \arg_{\theta}\min \frac{1}{N_s} \sum_{(\mathbf{x}, y) \in \mathcal{D}_s} -\log p_{\theta}(y|\mathbf{x})
\end{equation}

Where $p_{\theta}(\cdot|\mathbf{x})$ is the probability distribution on $C$ as predicted by the model in response to input $\mathbf{x}$. More specifically:

\begin{equation} \label{eq: prob}
    p(y=k|\mathbf{x}) = \frac{\exp{\langle\mathbf{w}_{k}, f_{\theta}(\mathbf{x})\rangle}}{\sum^{C}_{j=1}\exp{\langle\mathbf{w}_{j},f_{\theta}(\mathbf{x})\rangle}}
\end{equation}

$\langle \cdot \rangle$ refers to dot-product between features with class prototypes. 
As widely used in \cite{snell2017prototypical, qi2018low, chen2020new}, the novel class prototype $\mathbf{w}_c, c \in C$ is the mean feature from the support set $\mathcal{D}_s$:
\begin{equation}
    \mathbf{w}_c =\frac{1}{N_s} \sum_{\mathbf{x} \in \mathcal{D}_s}f_{\theta}(\mathbf{x}) \label{eq:mf}
\end{equation}

In our work, $f_{\theta}(\mathbf{x})$ is first pre-trained with meta-training set using cross-entropy loss; and further in each testing episode, $\theta^* = \theta^*(\mathcal{D}_s)$ is learned by finetuning $f_{\theta}(\mathbf{x})$ using $\mathcal{D}_s$. Given a test datum $\mathbf{x}$ where $(x, y) \in \mathcal{D}_q$, $y$ is predicted:

\begin{equation}
    \hat{y} = \arg \max_c p_{\theta*}(c|\mathbf{x})
\end{equation}

With this basic finetuning framework, we propose Distribution Calibration Module and Selected Sampling as introduced in the following section.

\subsection{Class-Agnostic Bias Reduction by Distribution Calibration Module} \label{sec:DCM}
We propose an easy-plug-in distribution calibration module(DCM) to reduce class-agnostic bias caused by domain difference. 

\textit{The first step to reduce class-agnostic bias is to calibrate skewed feature distribution.} A pre-trained feature extractor $f_{\theta}(\mathbf{x})$ could provide an initial feature space that general invariant features are learnt from a large-scale dataset.
$\theta^* = \theta^*(\mathcal{D}_{base})$ is sufficient to well classify those base classes which makes it inadequate to well distinguish novel classes. 
The overall feature distribution of novel classes may be skewed due to its domain property. And the feature distribution could be described statistically: 
\begin{equation}
    \mu = \frac{1}{N_s} \sum_{\mathbf{x}_i \in \mathcal{D}_s} f_{\theta}(\mathbf{x}_i), \sigma =  \frac{1}{N_s} \sum_{\mathbf{x}_i \in \mathcal{D}_s} (f_{\theta}(\mathbf{x}_i) - \mu)^2
\end{equation}

Note that $\mu$ and $\sigma$ are class-agnostic parameters describing the feature distribution for all obtained novel classes. For feature distributions that are skewed in some directions, the $\mu$ and $\sigma$ presented could be far from the normal distribution. We first apply to calibrate the distribution to approach zero centered mean and unit standard deviation: 
$\frac{f_i - \mathbf{\mu}}{\mathbf{\sigma}}$. This distribution calibration by feature normalization helps correct the skewed directions brought by large domain differences between base and novel classes. 

\begin{figure*}[t]
     \centering
     \begin{subfigure}{0.8\linewidth}
         \centering
         \includegraphics[width=\linewidth]{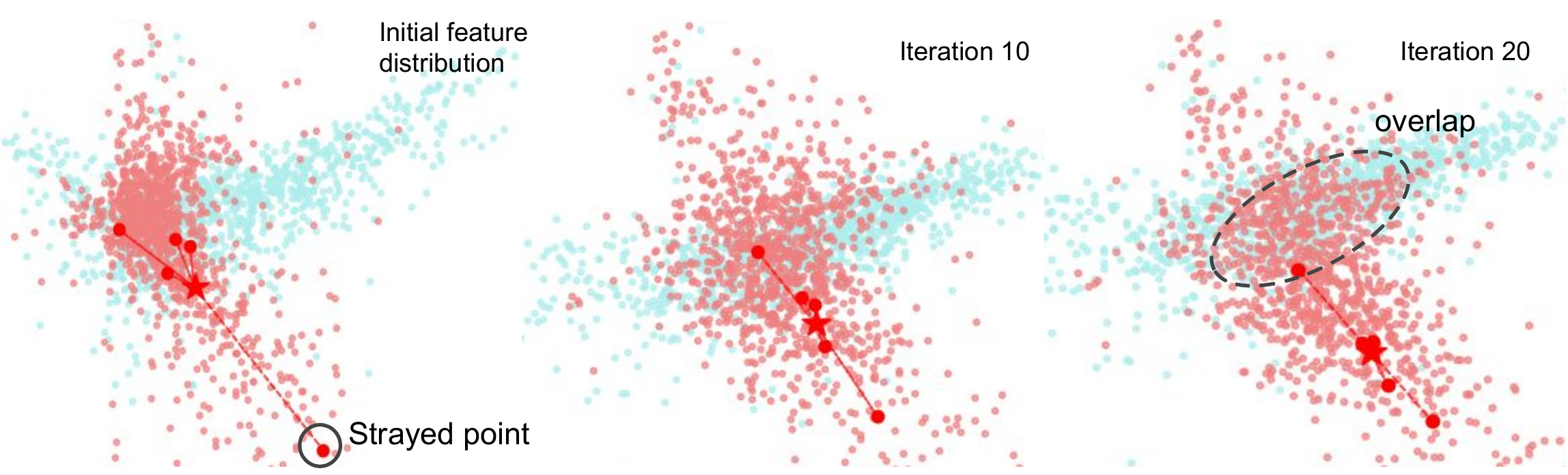}
         \caption{Without Selected Sampling. The optimization of feature distribution is distracted by a strayed point in the support set.
        }
         \label{finetune_aver}
     \end{subfigure}
     \begin{subfigure}{0.8\linewidth}
         \centering
         \includegraphics[width=\linewidth]{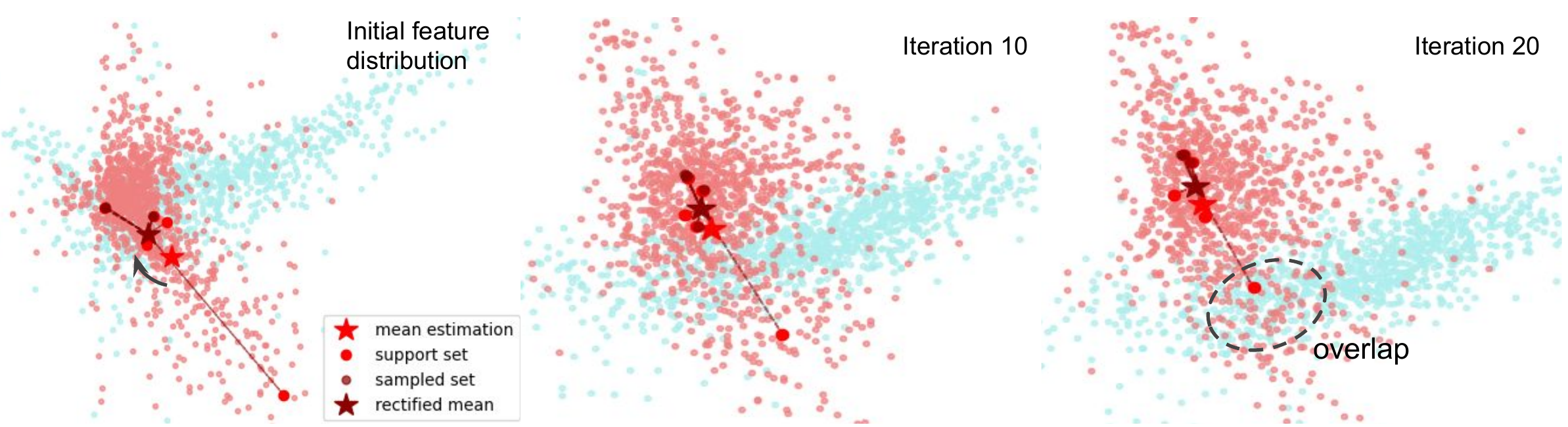}
         \caption{With Selected Sampling. SS helps to reduce bias in mean estimation brought by strayed points. And during optimization, the strayed points are gradually clustered towards the majority rather than distracting the clustering.
        }
         \label{finetune_rp}
     \end{subfigure}
    \caption{Finetuning w/o Selected Sampling on MNIST in 2d feature space. We plot all testing samples from the two novel classes for different iterations to visualize the change of feature distributions. 
Using Selected Sampling to reduce the bias in mean estimation, feature distributions are more compact within the same class. The overlap between the two classes is much smaller, with less sample density.}
    \label{fig: mnist}
    \vspace{-0.2cm}
\end{figure*}

\textit{Meanwhile, fast feature adaptation is enabled during finetuning.} For a feature vector, there are locations encoding class-sensitive information and locations encoding common information. Values on class-sensitive locations are expected to vary between classes to distinguish them. Similar values are obtained from common locations among all samples, representing some domain information but contributing little to classification. By this normalization, those locations that encode class-sensitive features are relatively stood out compared with the locations encoding common information. 
We further element-wisely multiply a scale vector among the calibrated feature embedding:
\begin{equation}
    \bar{f}_{\theta}(\mathbf{x}_i) = \frac{f_{\theta}(\mathbf{x}_i) - \mathbf{\mu}}{\mathbf{\sigma}} * \mathbf{s}
\end{equation}\label{scale}
$*$ is the element-wise multiplication. For simplicity, we use $\mathbf{f}_i =f_{\theta}(\mathbf{x}_i)$ for notation. The scale vector is learnable during fine-tuning. The element-wise multiplication allows each location on the scale vector could optimize independently, and thus the whole feature embedding could be reshaped on novel classes by this scale vector. 

As $\mathbf{s}$ is element-wisely multiplied with $f_i$, in the following discussion, we show only the partial derivative at a single location on the feature vector. In the 1-shot case with mean features as class prototypes, we have: 

\begin{equation}
    \frac{\partial L_i}{\partial s} \propto 
    \frac{f_i - \mu}{\sigma} [(p(y_i|x) - 1)\frac{f_i - \mu}{\sigma}
    + \sum^{C}_{j \neq y_i} p(j|x)\frac{f_j - \mu}{\sigma}] 
\end{equation}

After applying distribution calibration on $f_i$ and $f_j$, the gradients of $s$ for the class-sensitive locations have relatively larger values than the common ones. The difference between features is further addressed and enlarged correspondingly through gradient-based optimization. 
And the feature manifolds will fast adapt to the shape where distinguished parts are amplified.

\subsection{Class-Specific Bias Reduction by Selected Sampling}

\textit{Biased estimation in a class inevitably hinders the optimization of features.} The gradient for feature $f$ of $(\mathbf{x},y)$ during finetuning is:
\begin{equation}
    \frac{\partial L_i}{\partial \textbf{f}} = 
    (p(y|\textbf{x}) - 1)\mathbf{w}_y
    + \sum^{C}_{j \neq y} p(j|\textbf{x})\mathbf{w}_j
\end{equation}

As $p(y|\mathbf{x}) \leq 1$, the optimization of gradient descent focuses $f$ moving close towards the direction of $\mathbf{w}_y$, its ground-truth class prototypes. For a class $c$, mean feature from the support set is used as the class prototypes when computing the predicted probability \cite{snell2017prototypical, qi2018low, triantafillou2019meta}: $\mathbf{w}_c =\frac{1}{N_s} \sum_{\mathbf{x} \in \mathcal{D}_s}f_{\theta}(\mathbf{x})$. This is the empirical estimation of mean using the support set. We denote the true mean from the class distribution as $\mathbf{m}_c$. We further define the bias term $\delta_{c}$ between empirical estimation with its true value as: 
\begin{equation}\label{eq:bias}
    \delta_{c} = \mathbf{w}_{c} - \mathbf{m}_c
\end{equation}

For few-shot learning, as the $\mathbf{w}$ is estimated from a small number of data, $\delta_{c}$ is indeed not neglectful. 
As defined in Eq.~\ref{eq:bias}, $\mathbf{w}_{y}$ can be replaced by $\delta_{y} + \mathbf{m}_y$. Then the gradient of feature $f$ is:
\begin{equation}
    \frac{\partial L_i}{\partial \textbf{f}} = (p(y|\textbf{x}) - 1)\delta_{y} + (p(y|\textbf{x}) - 1)\mathbf{m}_y
    + \sum^{C}_{j \neq y} p(j|\textbf{x})\mathbf{w}_j
\end{equation}

The optimization of $f$ towards its class prototype $\mathbf{w}_y$ can be factorized into two parts: one part $(p(y|\textbf{x}) - 1)\delta_{y}$ dominated by the bias and the other by the true mean $\mathbf{m}_y$. Ideally, features are expected to tightly cluster around $\mathbf{m}$ for a refined feature distribution. However, $(p(y|\textbf{x}) - 1)\delta_{y}$ in the gradient distracts the optimization of $f$ by moving it close to the bias, which hinders its approaching to the true mean. And this inevitably impedes the optimization for few-shot learning. As shown in Fig.~\ref{finetune_aver}, points in the support set could be strayed from a majority in the class distribution. The strayed points enlarge bias in estimation, and thus during optimization, the clustering of the feature distribution is distracted by the bias. 

Augmenting more data is an efficient way to reduce bias. If more features could be sampled and added to the sequence of computing the class prototype within one class, effects caused by the bias will be vastly reduced. However, the feature distribution is unknown for each class, which disables the direct sampling from that distribution. 

\begin{figure}[t]
    \centering
    \includegraphics[width=1.0\linewidth]{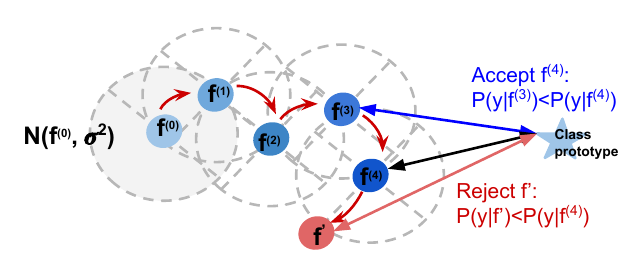}
    \caption{Illustration on Selective Sampling. With one feature $\mathbf{f}^{(0)}$ in the support set as an initial point, a new $\mathbf{f}^{(1)}$ is drawn from the proposal distribution $N(\mathbf{f}^{(0)}, \sigma^2)$; and once $\mathbf{f}^{(1)}$ is accepted, it becomes the next point for another sampling step. The sampling process will be terminated upon seeing one rejected point. The already sampled points $\mathbf{f}^{(1)}, \mathbf{f}^{(2)} ... \mathbf{f}^{(n)}$ will be appended to the support sets for updating the mean estimation. }
    \label{fig:proposal_ill}
\end{figure}{}

\textit{Without inference of actual feature distribution, we propose the Selected Sampling, which guides the Monte-Carlo Sampling under a proposal distribution.}
By taking advantage of each known data in the support set and let these a few samples guide the direction of a Monte-Carlo Sampling, we directly augment features into the support set.
For each known data point $(\mathbf{x}_i, y_i)$ the corresponding vector in the feature space is denoted as $\mathbf{f}_{i}$, a proposal distribution $Q(\mathbf{f}'|\mathbf{f}) = \mathcal{N}(\mathbf{f}_i, \Sigma)$ is used to sample $\mathbf{f}^{'}_{i}$.
$p(y|\mathbf{f})$ is a deterministic variable as the predicted logits from the classifier given a feature $\mathbf{f}$.
The sampled points are queried by the criterion $p(y_{i}|\mathbf{f}^{'}_{i}) > p(y_{i}|\mathbf{f}_{i})$ in determination of acceptance.
If accepted, $\mathbf{f}^{'}_{i}$ becomes the new starting feature point to run the next sampling step using proposal distribution $N(\mathbf{f}'_i, \sigma^2)$; if rejected, the sampling process for $(\mathbf{x}_i, y_i)$ terminates. We illustrate the sampling process in Fig.~\ref{fig:proposal_ill}. 

The proposal distribution ensures that samples are drawn from the vicinity around the known point during the process.  $\mathcal{N}(\mathbf{f}_i, \Sigma)$ is a multivariate Gaussian distribution centered with $\mathbf{f}_i$. The covariance matrix $\Sigma$ is an identity matrix scaled with a hyper-parameter $\sigma$,  which allows each location on features to be sampled independently. However, the proposal distribution is only a random walk process which brings no further constraints on the sampled points. 
With a feature $f = f_{\theta}(x)$, the acceptance criterion is whether the sampled feature will have a more significant predicted probability of belonging to the ground-true class or not, which is $p(y_{i}|\mathbf{f}^{'}_{i}) > p(y_{i}|\mathbf{f}_{i})$: 
\begin{equation}
    \frac{\exp{\langle\mathbf{w}_{k}, \mathbf{f}^{'}_{i}\rangle}}{\sum^{C}_{j=1}\exp{\langle\mathbf{w}_{j}, \mathbf{f}^{'}_{i}\rangle}} > \frac{\exp{\langle\mathbf{w}_{k}, \mathbf{f}_{i}\rangle}}{\sum^{C}_{j=1}\exp{\langle\mathbf{w}_{j}, \mathbf{f}_{i}\rangle}}
\end{equation}\label{eq: acceptance}
The numerator $\exp{\langle\mathbf{w}_{k}, \mathbf{f}^{'}_{i}\rangle}$ represents the distance between a feature with its class prototype, and the denominator $\sum^{C}_{j=1}\exp{\langle\mathbf{w}_{j}, \mathbf{f}^{'}_{i}\rangle}$ represents the overall distance between a feature with all class prototypes. This criterion indicates that a sampled point is accepted under the case either closer to its class prototype or further away from other classes in the high-dimensional feature space. Either way, the accepted point is ensured to provide helpful information that avoids the cons of random walk Sampling. 
This selected sampling on the feature space allows exploration of unknown feature space while still controlling the quality of sampling to optimize. 
As shown in Fig.~\ref{finetune_rp}, by enlarging the support set with selected samples, the bias in mean estimation is reduced. And Selected Sampling is an ongoing process for each iteration that helps to enhance the feature distribution clustering. 

\begin{figure*}
\begin{subfigure}{0.48\textwidth}
  \centering
  \includegraphics[width=1.0 \linewidth]{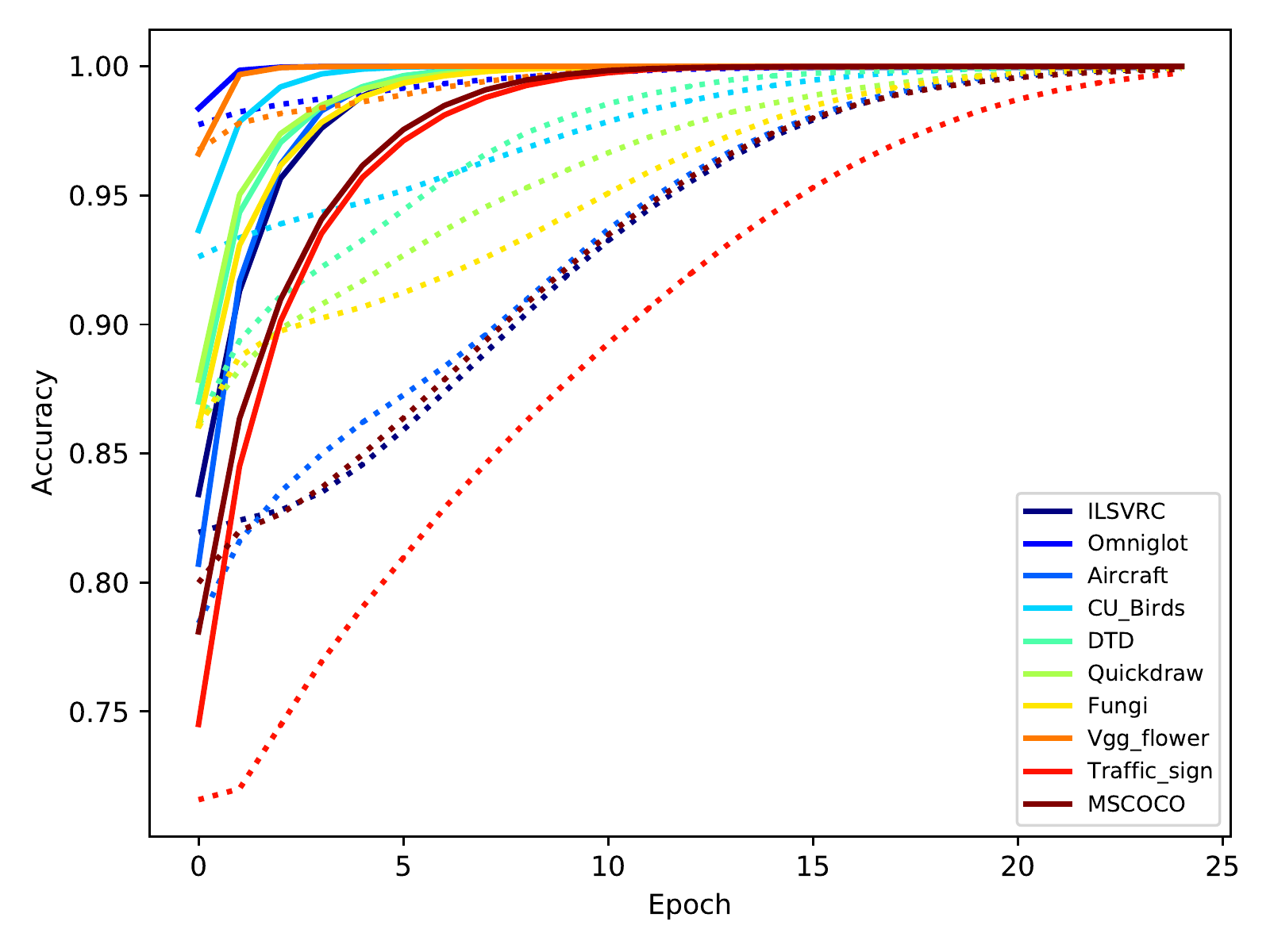}  
  \vspace{-0.2cm}
  \caption{Accuracy}
\end{subfigure}
\begin{subfigure}{0.48\textwidth}
  \centering
  \includegraphics[width=1.0 \linewidth]{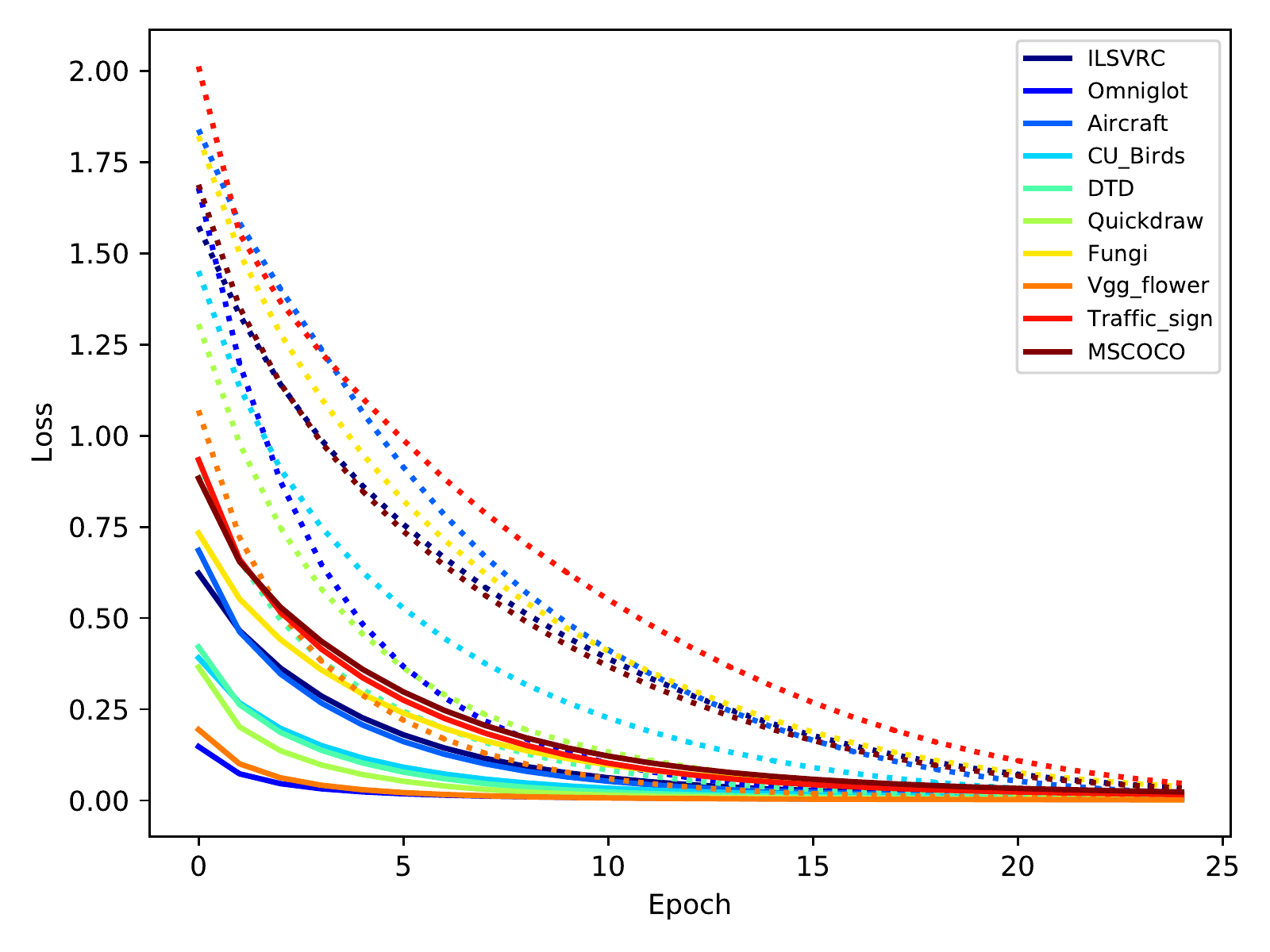}
  \vspace{-0.2cm}
  \caption{Loss}
\end{subfigure}
\caption{Convergence Curves on Meta-Dataset Standard Benchmark. For every dataset, we plot the average accuracy and loss over 600 episodes for each epoch during finetuning. The dashed curves represent finetuning backbone and the solid curves are for finetuning backbone with DCM. For all 10 datasets, finetuning backbone with DCM demonstrates faster speed to reach good training accuracy and much lower loss values. }
\label{fig:convergence}
\end{figure*}

\section{Experimental Validations} \label{sec:exp}
In this section, we first conduct comprehensive ablation experiments to verify the effectiveness of both DCM and SS and analyze how our method boosts performance under shot analysis. Then we compare our results with the other latest techniques. We follow the same setting and evaluation metrics in meta-Baseline \cite{chen2020new}. More specifically, when training the feature extractors, softmax cross-entropy loss is applied for learning. For finetuning on the (Meta-)test set, features and class prototypes are under normalization for the softmax loss. The temperature in the loss function is initialized to 10. 

We evaluate our method on Meta-Dataset\cite{triantafillou2019meta}, which is so far the most comprehensive benchmark for few-shot learning composed of multiple existing datasets in different domains.
The cross-domain property of meta-Dataset raises the real challenges for few-shot learning. To strictly verify the feature adaptation ability under different domains, we follow the evaluation by using the Imagenet-only training set to pre-train the feature extractor.

\subsection{Implementation Details}

\begin{table*}[!t]
    \centering
    \scalebox{0.98}{\begin{tabular}{c c c c|c c c c c c c c c c}
    \hline
     B & S & FN & SS & ILSVRC & Omni & Acraft & Birds & DTD & QDraw & Fungi & Flower & Sign & COCO \\
    \hline
     & & & & 58.47 & 69.80 & 54.35 & 76.51 & 75.47 & 77.68 & 44.47 & 89.10 & 48.18 & 56.93\\
     \checkmark & & & & 58.61 & 77.11 & 69.87 & 75.20 & 76.82 & 85.86 & 44.31 & 91.34 & 70.56 & 56.96 \\
     \checkmark & & \checkmark & & 59.10 & 77.69 & 71.84 & 78.80 & 76.48 & 86.00 & 47.28 & 92.30 & 74.43 & 57.94 \\
     \checkmark & \checkmark & \checkmark & & 59.96 & 78.70 & 72.32 & 78.30 & 76.96 & 86.04 & 47.51 & 91.95 & 76.39 & 
    57.32 \\
     \checkmark & \checkmark & \checkmark & \checkmark & \textbf{60.94} & \textbf{80.45} & \textbf{72.93} & \textbf{79.85} & \textbf{77.78} & \textbf{86.7} & \textbf{47.85} & \textbf{92.46} & \textbf{77.88} & \textbf{58.85}\\
    \hline
    \hline
    \end{tabular}}
    \caption{Ablation studies on DCM and SS using ResNet18. Results are reported using average of 600 episodes. Finetuning with Backbone(B), scale vector(S), feature normalization(FN) and selected sampling(SS) separately or combined are verified.  }
    \label{tab:ablation}

\end{table*}{}
\textbf{Pre-training the Backbone: Choice of the Network and Training Setting. } 

The ILSVRC-2012 \cite{ILSVRC15} in Meta-Dataset is splitted into 712 training, 158 validation and 130 test classes. We use the training set of 712 classes to train two feature extractors with backbones: ResNet18 and ResNet34. For ResNet18, we follow the same protocol in Meta-Baseline \cite{chen2020new}, which is: the images are randomly resized cropped to 128x128, horizontal flipped and normalized. For ResNet34, we follow the same structure modification in \cite{doersch2020crosstransformers} which uses stride 1 and dilated convolution for the last residual block and the input image size is 224x224. The initial learning rate is set to 0.1 with 0.0001 weight decay and decreases by a factor of 0.1 every 30 epochs with total 90 epochs. Both models are trained using the SGD optimizer with batch size 256.

\textbf{Setting of Evaluation and Fine-tuning.}
The general evaluation on Meta-Dataset utilizes a flexible sampling of episodes \cite{triantafillou2019meta}, which allows a maximum of 500 images in the support set in one episode. In the finetuning stage, the scale vector $s$ is initialized with value $1$. Data argumentation works as resizing and center cropping images to 128x128(ResNet18) and 224x224(ResNet34) followed by normalization. We follow the setting in \cite{dhillon2019baseline} as described with learning rate of 0.00005, Adam optimizer and 25 total epochs. $\sigma$ in the proposal distribution for sampling is set to $0.1$. Finetuning experiments are conducted with whole batch update.

\subsection{Ablation Studies}
We first study the importance of applying DCM during finetuning with the typical class prototype(mean feature from support set) and then upon applying DCM we add SS to rectify class prototypes. All ablation results are in Table.~\ref{tab:ablation}.

\textbf{DCM enables competitive domain-agnostic fast feature adaptation for finetuning.} 
There are two functionalities in DCM: feature normalization and the scale vector multiplication. We first independently evaluate the performance gain brought by feature normalization(FN) and then also verify the importance of the scale vector(S) in fast feature adaptation. 
Only finetuning the backbone cannot guarantee the performance improvement over all datasets due to the different domain gaps, especially for CU-Birds and Fungi as shown in Table.~\ref{tab:ablation}. 
Among these datasets where finetuning the backbone is not practically working, the performances improve by $3.10\%$ on CU-Birds and $3.39\%$ on Fungi by simply adding feature normalization. And by adding the scale vector, the performance is further improved over 7 out of 10 datasets. These results indicate that DCM improves the generalization of finetuning over datasets from different domains. We further plot the training losses and accuracy during the finetuning iterations in Figure.~\ref{fig:convergence}. The plot shows that finetuning with DCM owes supreme convergence speed compared with direct finetuning. The advantage in convergence speed demonstrates the property of fast feature adaptation of DCM.

\begin{table*}[!t]
    \centering
    \scalebox{0.82}{\begin{tabular}{c|c|c|c|c|c|c|c|c|c|c|c}
    \hline
    \hline
    Method  & Backbone & ILSVRC & Omni & Acraft & Birds & DTD & QDraw & Fungi & Flower & Sign & COCO \\
    \hline
    \hline
    fo-Proto-MAML
    & \multirow{2}{*}{-} & \multirow{2}{*}{49.53} & \multirow{2}{*}{59.98} & \multirow{2}{*}{53.10} & \multirow{2}{*}{68.79} & \multirow{2}{*}{66.56} & 
    \multirow{2}{*}{48.96} & 
    \multirow{2}{*}{39.71} & 
    \multirow{2}{*}{85.27} & 
    \multirow{2}{*}{47.12} & \multirow{2}{*}{41.00}\\
    \cite{triantafillou2019meta} &  &  &  &  &  &  &  &  &  &  & \\
    CNAPS & \multirow{2}{*}{-} & \multirow{2}{*}{50.60} & \multirow{2}{*}{45.20} & \multirow{2}{*}{36.00} & \multirow{2}{*}{60.7} & \multirow{2}{*}{67.5} & \multirow{2}{*}{42.3} & \multirow{2}{*}{30.1} & \multirow{2}{*}{70.7} & \multirow{2}{*}{53.3} & \multirow{2}{*}{45.2}\\
    \cite{requeima2019fast} &  &  &  &  &  &  &  &  &  &  & \\
    BOHB-Ensemble & \multirow{2}{*}{-} & \multirow{2}{*}{55.39} & \multirow{2}{*}{77.45} & \multirow{2}{*}{60.85} & \multirow{2}{*}{73.56} & \multirow{2}{*}{72.86} & \multirow{2}{*}{61.16} & \multirow{2}{*}{44.54} & \multirow{2}{*}{90.62} & \multirow{2}{*}{57.53} & \multirow{2}{*}{51.86}\\
    \cite{saikia2020optimized} &  &  &  &  &  &  &  &  &  &  & \\
    LR & \multirow{2}{*}{ResNet18} & \multirow{2}{*}{60.14} & \multirow{2}{*}{64.92} & \multirow{2}{*}{63.12} & \multirow{2}{*}{77.69} & \multirow{2}{*}{78.59} & \multirow{2}{*}{62.48} & \multirow{2}{*}{47.12} & \multirow{2}{*}{91.60} & \multirow{2}{*}{77.51} & \multirow{2}{*}{57.00}\\
    \cite{tian2020rethinking} &  &  &  &  &  &  &  &  &  &  & \\
    Meta-Baseline & \multirow{2}{*}{ResNet18} & \multirow{2}{*}{59.20} & \multirow{2}{*}{69.10} & \multirow{2}{*}{54.10} & \multirow{2}{*}{77.30} & \multirow{2}{*}{76.00} & \multirow{2}{*}{57.30} & \multirow{2}{*}{45.40} & \multirow{2}{*}{89.60} & \multirow{2}{*}{66.20} & \multirow{2}{*}{55.70} \\
    \cite{chen2020new}&  &  &  &  &  &  &  &  &  &  & \\
    Transductive-finetuning & \multirow{2}{*}{WRN-28-10} & \multirow{2}{*}{60.53} & \multirow{2}{*}{82.07} & \multirow{2}{*}{72.40} & \multirow{2}{*}{82.05} & \multirow{2}{*}{\textbf{80.47}} & \multirow{2}{*}{57.36} & \multirow{2}{*}{47.72} & \multirow{2}{*}{92.01} & \multirow{2}{*}{64.37} & \multirow{2}{*}{42.86} \\
    \cite{dhillon2019baseline}&  &  &  &  &  &  &  &  &  &  & \\
    CTX-best & \multirow{2}{*}{ResNet34} & \multirow{2}{*}{62.76} & \multirow{2}{*}{\textbf{82.21}} & \multirow{2}{*}{79.49} & \multirow{2}{*}{80.63} & \multirow{2}{*}{75.57} & \multirow{2}{*}{72.68} & \multirow{2}{*}{\textbf{51.58}}  & \multirow{2}{*}{\textbf{95.34}} & \multirow{2}{*}{82.65} & \multirow{2}{*}{\textbf{59.90}} \\
    \cite{doersch2020crosstransformers}&  &  &  &  &  &  &  &  &  &  & \\
    \hline
    Classifier-Baseline & ResNet18 & 58.47 & 69.80 & 54.35 & 76.51 & 75.47 & 77.68 & 44.47 & 89.10 & 48.18 & 56.93\\
    DCM+SS & ResNet18 & 60.94 & 80.45 & 72.93 & 79.85 & 77.78 & 86.7 & 47.85 & 92.46 & 77.88 & 58.85\\
    Classifier-Baseline & ResNet34 & 60.37 & 72.38 & 61.19 & 77.93 & 75.91 & 79.76 & 42.77 & 89.80 & 48.56 & 51.79\\
    DCM+SS & ResNet34 & \textbf{64.58} & 81.77 & \textbf{79.67} & \textbf{84.94} & 77.89 & \textbf{87.14} & 49.34 & 93.24 & \textbf{88.65} & 57.69 \\
    \hline\hline
    \end{tabular}}
    \caption{Results on Standard Benchmark of Meta-Dataset. We provide the statistical results of over 600 episodes. As shown above, our method brings consistent performance improvements over all ten datasets compared with recent works. This demonstrates our proposed bias reduction methods could effectively work for different data domains. }
    \label{tab:sota}

\end{table*}{}

\textbf{Selected Sampling can consistently improve performance over all datasets. }
With adding selected sampling, performance on all datasets is improved from $0.34\%$ to $1.75\%$. And for 6 out of 10 datasets, the performance is boosted by roughly $1\%$. Especially for ILSVRC, Birds, and MSCOCO, the performance gains brought by SS are the most significant compared with finetuning backbone or adding DCM with backbone. These datasets are diverse in objects and cover significant variations within one class. For example, CU-Birds requires high demands on fine-grained classification. The performance gain strongly indicates that using Selected Sampling to rectify class prototypes works well with features in different domains. 

\begin{figure}[t]
    \centering
    \includegraphics[width=1.0\linewidth]{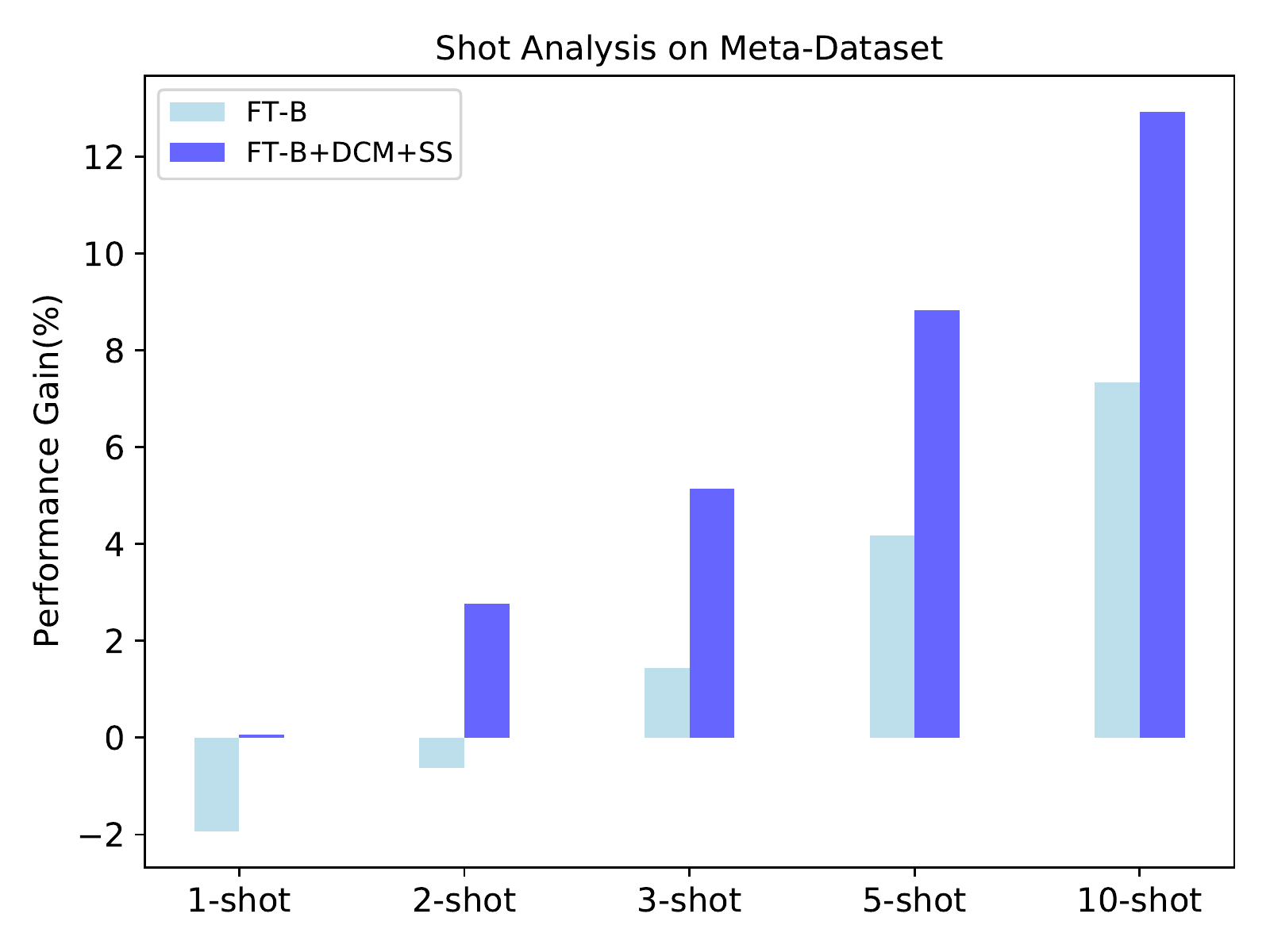}
    \caption{Performance Gain using Finetuning w/o DCM+SS under Different Number of Shots. Performance is averaged over 600 episodes and reported performance gains are averaged over different datasets in Meta-Dataset. }
    \label{fig:shots}
      \vspace{-0.5cm}
\end{figure}

\textbf{DCM with SS plays an essential role under extreme few shots.}
Meanwhile, we further evaluate how DCM+SS powers finetuning, especially under extreme few shots. As shown in Figure.~\ref{fig:shots}, by fixing the number of shots in one episode, we provide the average performance gain by running 600 episodes for each dataset. 
Finetuning the backbone leads to a performance drop with only one or two shots per class, while DCM+SS can dramatically compensate for the performance loss.
Meanwhile, by only increasing the number of shots(comparing FT-B performance on 2-shot and 3-shot), finetuning can be improved by a relatively small range. Adding DCM+SS leads to much more performance gain simply on 2-shot cases. DCM+SS essentially boosts performance on a few shots. In all, DCM+SS shows consistent significant performance increases over over extreme few shot cases. 

\textbf{DCM with SS powers finetuning. } 
From Table.~\ref{tab:ablation}, finetuning improves the performance around $7.31\%$ to $15.52\%$ on several datasets, but performance drops on Birds and Fungi. DCM+SS with finetuning the backbone shows consistent performance boosts on all datasets from $1.92\%$(mscoco) to $29.7\%$(Traffic sign). We hope this results encourage further explorations on bias reduction in feature distribution. 

\subsection{Compare with the State of the Art Performance}
We report our results under different backbone models and provide a comparison over other popular methods in Table.~\ref{tab:sota}. Compared with directly using the pre-trained feature extractor for evaluation, finetuning backbone with DCM+SS boosts performance significantly over all datasets. The results are consistent among both backbones. This demonstrates the effectiveness of directly applying finetuning using the support set. Comparing the performance on ResNet18 and ResNet34, we first observe that a larger backbone with a larger input image size gives a better quality of the feature extractor. Furthermore, our method shows the adaptation ability of the pre-trained feature extractor to new data domains can even be improved when the feature extractor itself is more powerful. 
By a simple testing-time finetuning, we achieve the State-of-the-Art performance over several datasets and closely competitive results for all datasets with ResNet18 and ResNet34. \cite{tian2020rethinking} only finetunes a classifier on the testing set, and with the same backbone ResNet18, our method surpasses its results with a large margin on most datasets. This addresses the importance of refining novel-class features for better generalization.  
For \cite{doersch2020crosstransformers} which overpasses our results on four datasets, besides a pre-trained feature extractor, a comprehensive meta-training process using seven days to converge as reported in their work is also utilized. The ResNet34 feature extractor we use is only trained by supervised classification loss using the training set. Meanwhile, our method is computationally efficient as we barely involve any network structure change(only one DCM layer with a scale vector). The sampling is conducted in an efficient full batch style. \cite{dhillon2019baseline} includes extra query set during transductive finetuning, which leads to a better result on DTD. While our finetuning is only using the support set. And we surpass the results on the other nine datasets and have a consistent performance gain on DTD.

\section{Conclusion}

We show in our experiments that without any meta-training process, the fast feature adaptation can also be achieved by better understanding biases in feature distribution for few-shot learning. 
We hope our work could provide insight into the importance of bias reduction in distribution when dealing with datasets from different domains for few-shot learning. 


\end{document}